\documentclass{article}
\usepackage{spconf,amsmath,graphicx,hyperref}
\usepackage{booktabs}
\usepackage{algorithm,algpseudocode}
\graphicspath{{figures/}}

\newcommand{\epsu}{\epsilon_u}
\newcommand{\epsi}{\epsilon_i}
\newcommand{\epst}{\epsilon_t}
\newcommand{\method}{Attention-Scoped Guidance}
\newcommand{\asg}{ASG}

\title{Attention-Scoped Guidance: Training-Free Spatial Control for Image Editing}
\name{Zeyan Li$^1$, Wei Zhou$^2$, Hadi Amirpour$^3$, Minghao Zou$^2$, Panqi Yang$^4$, Jianfeng Xu$^1$\sthanks{Corresponding author.}}
\address{$^1$Shanghai Jiao Tong University, $^2$Cardiff University, \\ 
$^3$University of Klagenfurt, $^4$Xi'an Jiaotong University}

\begin{document}
\ninept
\maketitle

\begin{abstract}
Instruction-guided image editing should change what the instruction names and leave the rest of the image untouched. In dual classifier-free guidance (CFG), an editor combines two directions at every denoising step, one that pushes toward the instructed edit and one that pulls back toward the source image, using global weights. We introduce \method{} (\asg{}), a sampler wrapper that makes these weights spatial. It reads a soft support map from the instruction attention that the editor already computes, then weakens text guidance where support is low and strengthens image anchoring where support is high. The wrapper requires no training, no external mask, and no additional network evaluation. On the full MagicBrush and PIE-Bench++ splits, \asg{} improves preservation-oriented metrics, leading three of four MagicBrush metrics and PIE-Bench++ background PSNR. A dose-matched control that removes the spatial placement loses up to 0.73 CLIP on PIE-Bench++, confirming that the spatial allocation itself carries the gain.
\end{abstract}


\section{Introduction}

\begin{figure}[t]
\centering
\includegraphics[width=\columnwidth]{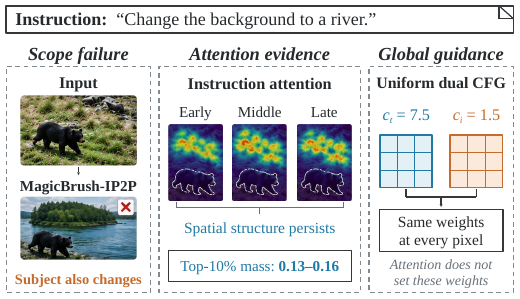}
\caption{An illustrative failure from a ten-case probe, motivating \method{}. Left: for a background-only MagicBrush instruction, the underlying editor changes both the scene and the subject. Center: the displayed attention maps retain spatial structure through denoising, with nearly unchanged top-10\% mass. Right: the dual-CFG combination uses global text and image weights, motivating explicit spatial allocation.}
\label{fig:discovery}
\end{figure}

Instruction-guided image editing should change the content named by the instruction and leave the rest of the image untouched. Diffusion editors such as InstructPix2Pix (IP2P) \cite{brooks2023instructpix2pix} often violate this boundary. A local replacement regenerates the scene, changes the subject's identity, or modifies an untouched region. MagicBrush makes this tension measurable through paired source, target, and instruction data \cite{zhang2023magicbrush}. Fig.~\ref{fig:discovery} shows the characteristic failure, where the background changes as requested but the subject changes with it.

Localizing an edit to the content the instruction names is therefore the central difficulty. Existing editors pursue this either by changing what the editor learns, through human feedback, task-specific finetuning, or broader instruction data \cite{zhang2024hive,geng2024instructdiffusion,fu2024mgie,chen2025instructclip}, or by constraining sampling with attention injection, inversion, or region blending \cite{hertz2023prompttoprompt,mokady2023nulltext,avrahami2023blended}. Both lines improve how faithfully an instruction is followed or how much structure survives, but the guidance weights themselves remain global scalars.

The methods closest to this problem control locality explicitly. Grounded variants obtain a region from an external detector \cite{shagidanov2024grounded}, ZONE derives one zero-shot from internal attention \cite{li2024zone}, Focus on Your Instruction combines cross-attention modulation with mask-guided disentangled sampling \cite{guo2024focus}, and LIME regularizes attention to keep edits local \cite{simsar2025lime}; in text-to-image generation, S-CFG makes guidance weights spatially varying across semantic regions \cite{shen2024scfg}. The underlying attention maps are known to carry spatial semantics \cite{tang2023daam,chefer2023attend}. S-CFG spatially rescales a single text-guidance weight for text-to-image generation; our focus is the dual-CFG combination in instruction-guided editing, where the edit is committed by two global scalars, the same weight at every pixel. Our ten-case probe confirms the mismatch: instruction attention remains spatially structured through denoising in successful and failed edits alike, and concentrated maps appear in failures as often as in successes. The spatial signal is present, but the sampler never couples it to the edit, so a locally supported instruction leaks through the uniformly weighted text direction into unrelated pixels.

\begin{figure*}[ht]
\centering
\includegraphics[width=0.9\textwidth]{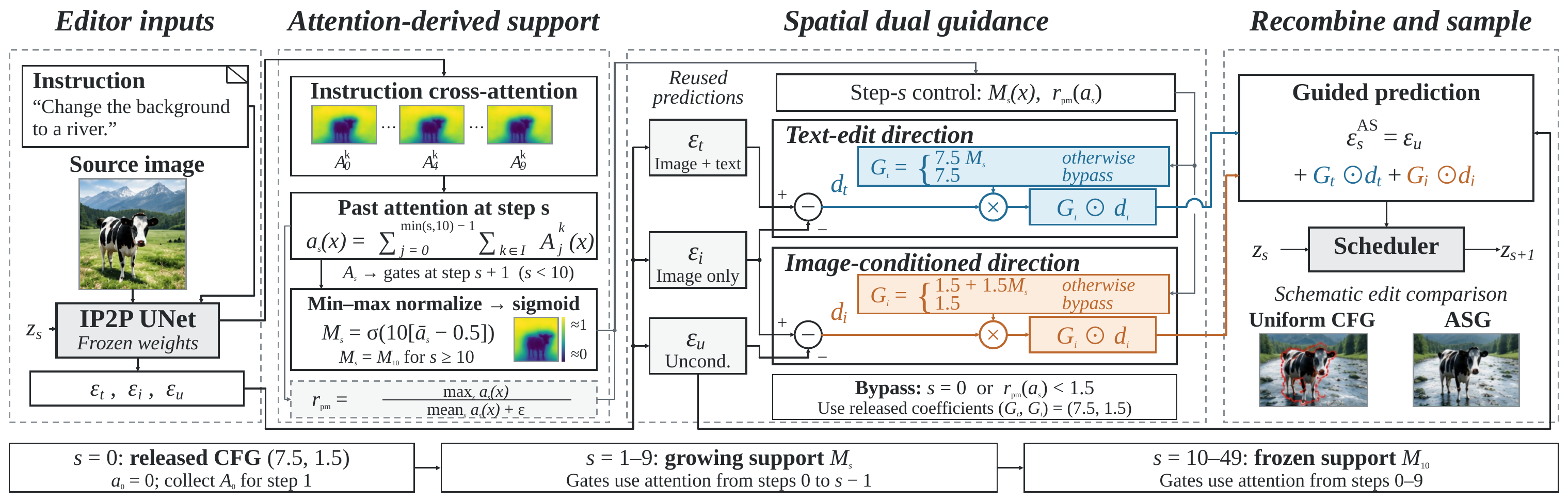}
\caption{Method overview, left to right. Given a source image and an instruction, the frozen editor evaluates three noise predictions at every denoising step. Instruction cross-attention, averaged over heads and layers and summed over tokens and the first ten steps, yields a soft support $M(x)$. From the second step on, this support makes the two CFG weights spatial: the text-edit direction is weakened where $M$ is low, and the image anchor is strengthened where $M$ is high; the support freezes after step ten. The reweighted directions are recombined and sampled unchanged, completing the edit within the original sampling trajectory, with no additional network evaluations.}
\label{fig:method}
\end{figure*}

We therefore propose \method{} (\asg{}), illustrated in Fig.~\ref{fig:method}. ASG reads one soft support from the instruction attention that the editor already computes and uses it to allocate both guidance residuals: the text-edit direction is weakened where support is low, and image anchoring is strengthened where support is high. Spatially varying guidance and attention-derived regions are established ideas \cite{shen2024scfg,guo2024focus}; the distinct element is the allocation rule at the dual-CFG combination itself, where a single support assigns different coefficients to the two residuals. A peak-to-mean bypass restores the released sampler whenever attention is diffuse.

Experiments on both public benchmarks quantify the benefit. \asg{} leads three of four MagicBrush metrics and PIE-Bench++ background PSNR against seven released editors, while DIM remains stronger on PIE-Bench++ semantic CLIP, so the method occupies the preservation end of the trade-off. A dose-matched control attributes the gain to the spatial allocation itself, and the dose analysis shows that stronger anchoring begins to resist requested deletions. Our contributions are a training-free sampler wrapper that makes the dual-CFG weights spatial without external input, full-population evidence of its preservation-oriented improvement, and controls and dose analyses that attribute the gain and bound the anchor strength.

\section{Spatial Allocation of Dual Guidance}
\label{sec:method}

IP2P builds on latent diffusion \cite{rombach2022latent} and classifier-free guidance \cite{ho2021classifier}. At denoising step $s$, it evaluates three predictions: an unconditioned prediction $\epsu(x)$, an image-conditioned prediction $\epsi(x)$, and an instruction-conditioned prediction $\epst(x)$. These are combined as
\begin{equation}
 \epsilon_s(x)=\epsu(x)+c_t[\epst(x)-\epsi(x)]
 +c_i[\epsi(x)-\epsu(x)],
 \label{eq:cfg}
\end{equation}
where the official weights are $c_t=7.5$ and $c_i=1.5$. The first difference is the text-edit direction and the second is the image-identity direction,
\begin{align}
 d_t(x)&=\epst(x)-\epsi(x), &
 d_i(x)&=\epsi(x)-\epsu(x).
 \label{eq:directions}
\end{align}
The direction $d_t$ introduces the content required by the instruction, while $d_i$ pulls the sample back toward the source image. Equation~\eqref{eq:cfg} assigns each direction its own global scalar, so every spatial location receives the same mixture of change and stay. \asg{} replaces these scalars with spatial weights derived from existing attention. The residuals already carry spatial information, and the intervention adds explicit spatial control over their relative contributions.

\subsection{Attention-Derived Support}

Let $A_s^k(x)$ denote cross-attention from spatial location $x$ to instruction token $k$, and let $\mathcal I$ contain the non-special tokens of the edit instruction. We read $A$ from the cross-attention layers at half the latent resolution, averaged over heads and layers. The support available at step $s$ accumulates only the attention captured before it and freezes after step ten:
\begin{align}
 a_s(x)&=\sum_{j=0}^{\min(s,10)-1}\sum_{k\in\mathcal I}A_j^k(x), \quad a_0(x)=0, \\
 \bar a(x)&=\frac{a(x)-\min_x a(x)}{\max_x a(x)-\min_x a(x)+\varepsilon}, \\
 M(x)&=\sigma\!\left(10[\bar a(x)-0.5]\right),
 \label{eq:support}
\end{align}
where $\varepsilon=10^{-6}$. At step $s$ the gates use $a_s$; we omit the step subscript below for brevity. Min-max normalization places each image's attention on a common scale, and the sigmoid provides a soft transition between low and high responses, so $M$ is a soft support rather than a binary mask. We measure the map's concentration by
\begin{equation}
 r_{\rm pm}=\frac{\max_x a(x)}{|\Omega|^{-1}\sum_{x\in\Omega}a(x)+\varepsilon},
 \label{eq:peakmean}
\end{equation}
where $\Omega$ is the spatial domain of the attention map. If $r_{\rm pm}<1.5$, attention is diffuse and carries no usable spatial signal: \asg{} is bypassed, and the released coefficients $G_t(x)=7.5$ and $G_i(x)=1.5$ apply at every pixel. The spatial rules below apply when this bypass is inactive. The threshold $r_{\rm pm}<1.5$ and the sigmoid scale $10$ were selected on the development split.

Accumulating over several steps averages out the noise of any single step's attention; early steps decide the global layout while later steps refine local detail, so the support is read where the layout is decided and held fixed before texture forms. The peak-to-mean bypass keeps the method conservative: it intervenes only where attention concentrates.

\begin{table*}[t]
\centering
\caption{Main full-population comparison against released external editors. MagicBrush-I and MagicBrush-T denote the independent and iterative MagicBrush test protocols. Best values within each metric are bold. The last row reports \method{}.}
\label{tab:main}
\footnotesize
\setlength{\tabcolsep}{3.8pt}
\renewcommand{\arraystretch}{1.12}
\begin{tabular*}{\textwidth}{@{\extracolsep{\fill}}lccccccc@{}}
\toprule
& \multicolumn{2}{c}{MagicBrush-I} &
\multicolumn{2}{c}{MagicBrush-T} &
\multicolumn{3}{c}{PIE-Bench++} \\
\cmidrule(lr){2-3}\cmidrule(lr){4-5}\cmidrule(l){6-8}
System & CLIP-I$\uparrow$ & L1$\downarrow$ & CLIP-I$\uparrow$ & L1$\downarrow$ &
CLIP-W$\uparrow$ & CLIP-E$\uparrow$ & PSNR-BG$\uparrow$ \\
\midrule
InstructPix2Pix & 0.8509 & 0.1135 & 0.8099 & 0.1491 & 24.001 & 23.457 & 80.656 \\
MagicBrush-IP2P & 0.9177 & 0.0741 & 0.8849 & 0.1042 & 23.612 & 22.985 & 81.388 \\
InstructDiffusion & 0.9228 & 0.0706 & 0.8912 & 0.0995 & 23.903 & 23.155 & 81.591 \\
MGIE & 0.9087 & 0.0812 & 0.8745 & 0.1133 & 21.357 & 21.080 & 79.461 \\
InstructCLIP & 0.8794 & 0.0962 & 0.8360 & 0.1291 & 23.441 & 22.767 & 80.985 \\
DIM & 0.9256 & 0.0698 & \textbf{0.8981} & 0.0973 & \textbf{26.745} & \textbf{25.933} & 81.023 \\
Kontinuous Kontext & 0.9099 & 0.0814 & 0.8724 & 0.1150 & 25.511 & 24.718 & 81.220 \\
\midrule
\textbf{\method{}} & \textbf{0.9296} & \textbf{0.0624} & 0.8961 & \textbf{0.0901} &
23.568 & 22.858 & \textbf{82.173} \\
\bottomrule
\end{tabular*}
\end{table*}

\subsection{Spatial Dual Guidance}

The two global scalars become two spatial weights. The text-edit direction is attenuated where support is low,
\begin{equation}
 G_t(x)=g_{\rm lo}+(7.5-g_{\rm lo})M(x),
 \label{eq:textgate}
\end{equation}
where $g_{\rm lo}=0$ by default, and the image-conditioned contribution is increased where support is high,
\begin{equation}
 G_i(x)=1.5+(g_{\rm hi}-1.5)M(x).
 \label{eq:imagegate}
\end{equation}
The two weights serve different spatial roles because text and image guidance act on different CFG directions. We set $g_{\rm lo}=0$ so that text guidance is nearly suppressed outside the support. With this default, the selected dose $g_{\rm hi}=3.0$ gives a pixel with $M(x)=0.5$ the coefficients $G_t=3.75$ and $G_i=2.25$, that is, half-strength edit pressure plus a moderately reinforced anchor. The continuous weights keep the transition zone between the instructed region and its surroundings free of visible seams.

\subsection{Recombine and Sample}

The gated directions are recombined with the unconditioned prediction,
\begin{equation}
 \epsilon^{\rm AS}_s(x)=\epsu(x)+G_t(x)d_t(x)+G_i(x)d_i(x).
 \label{eq:asg}
\end{equation}
Subtracting Eq.~\eqref{eq:cfg} makes the intervention explicit,
\begin{equation}
 \Delta\epsilon_s(x)=-7.5[1-M(x)]d_t(x)
 +(g_{\rm hi}-1.5)M(x)d_i(x).
 \label{eq:delta}
\end{equation}
At the idealized endpoint $M(x)=0$ the text-edit direction is canceled while the default image anchor is kept; at $M(x)=1$ the original text guidance is retained and only the image anchor is increased, and intermediate values interpolate continuously. The original sampler is recovered by
\begin{equation}
 M\equiv1,\quad g_{\rm hi}=1.5
 \quad\Longrightarrow\quad
 \epsilon^{\rm AS}_s=\epsilon_s.
 \label{eq:identity}
\end{equation}
When $M\equiv1$, $G_t=7.5$ for any $g_{\rm lo}$, so the identity holds regardless of the lower bound. The dose study in Table~\ref{tab:dosefull} selects $g_{\rm hi}=3.0$ from $\{2.25,3.0,4.5\}$ using the development similarity metrics and failure tail. Algorithm~\ref{alg:asg} assembles one full edit.

\begin{algorithm}[ht]
\caption{Attention-scoped dual-CFG sampling for one edit.}
\label{alg:asg}
\begin{algorithmic}[1]
\Require source latent, instruction tokens $\mathcal I$, anchor dose $g_{\rm hi}$
\State $a(\cdot)\gets 0$
\For{$s=0$ \textbf{to} $49$}
  \If{$s=0$ \textbf{or} $r_{\rm pm}(a)<1.5$} \Comment{$a$: steps $<s$}
    \State $(G_t,G_i)\equiv(7.5,1.5)$ \Comment{bypass: released coefficients}
  \Else
    \State $G_t(x)\gets g_{\rm lo}+(7.5-g_{\rm lo})M(x)$;\; $G_i(x)\gets 1.5+(g_{\rm hi}-1.5)M(x)$, with $M$ built from $a$ (Eq.~\eqref{eq:support})
  \EndIf
  \State evaluate $\epsu,\epsi,\epst$ as in IP2P (Eq.~\eqref{eq:cfg}) and update the latent with $\epsilon^{\rm AS}_s(x)=\epsu+G_t d_t+G_i d_i$ (Eq.~\eqref{eq:asg})
  \If{$s<10$}
    \State $a(x)\gets a(x)+\sum_{k\in\mathcal I}A_s^k(x)$ \Comment{gates step $s{+}1$}
  \EndIf
\EndFor
\end{algorithmic}
\end{algorithm}

\section{Experiments}
\label{sec:experiments}

\subsection{Evaluation Protocol}

We use the full public splits of MagicBrush and PIE-Bench++ \cite{ju2024pnpinversion,piebenchpp}. PIE-Bench++ retains the 700 PIE-Bench images with revised prompts and annotations; all systems, including the released baselines, are evaluated on this same distribution under one fixed generation seed, so every comparison is matched. MagicBrush has 1,053 turn-level examples in each independent and iterative test protocol, and PIE-Bench++ has 700 examples. We use PIE-Bench++ rather than EditVal \cite{basu2023editval} because it supplies both edit and background regions. Background PSNR zeroes the official edit region before comparing; global edits whose mask covers the full image leave no background and return the evaluator's $100$\,dB cap, so discrimination comes from the $171$ local edits. Mechanism tests use all 528 MagicBrush development turns with the official pixel-L1 and CLIP-I ViT-B/32 evaluators and the released preprocessing. The CLIP backbone follows \cite{radford2021clip}.

The no-op reference is the source image itself. We use ``catastrophic'' as shorthand for a CLIP-I gap to this no-op reference below $-0.10$. Every edit uses the same $g_{\rm hi}$, and no benchmark mask enters the sampler.

\subsection{The Preservation--Editability Trade-off}

Table~\ref{tab:main} compares \asg{} with seven released editors \cite{brooks2023instructpix2pix,zhang2023magicbrush,geng2024instructdiffusion,fu2024mgie,chen2025instructclip,zeng2026dim,parihar2026kontinuous} on both benchmarks. \asg{} leads both MagicBrush independent metrics (0.9296/0.0624) and iterative L1 (0.0901), while DIM leads iterative CLIP-I (0.8981 versus 0.8961), and DIM and Kontinuous Kontext lead both PIE-Bench++ semantic CLIP metrics; \asg{} gives the highest background PSNR (82.173). The two strongest semantic baselines, DIM and Kontinuous Kontext, exceed \asg{} on both PIE-Bench++ CLIP scores but trail it by 1.15 and 0.95\,dB of background PSNR, widening to 4.7 and 3.9\,dB on the 171 local edits with nonempty background.

The complete system exceeds base IP2P by 0.0787 CLIP-I and 0.0511 L1 on the independent protocol. The harder margin is against MagicBrush-IP2P, the checkpoint already finetuned for this benchmark, where the gains are 0.0119/0.0117 independently and 0.0112/0.0141 iteratively, from sampling alone. On PIE-Bench++, the background-PSNR lead over the best baseline (InstructDiffusion, 81.591) is 0.582 dB.

\begin{table}[t]
\centering
\caption{Branch ablations on the 528-turn MagicBrush development split. ``Base'' is the public IP2P checkpoint; ``incumbent'' is the finetuned backbone used for the main results. ``Cat.'' counts turns whose CLIP-I gap to the no-op reference falls below $-0.10$.}
\label{tab:ablation}
\scriptsize
\setlength{\tabcolsep}{3pt}
\begin{tabular*}{\columnwidth}{@{\extracolsep{\fill}}lccc@{}}
\toprule
Configuration & CLIP-I$\uparrow$ & L1$\downarrow$ & Cat.$\downarrow$ \\
\midrule
Base: uniform CFG & 0.8625 & 0.1104 & 152 \\
Base: text gating only & 0.9303 & 0.0653 & 12 \\
\midrule
Incumbent: uniform CFG & 0.9217 & 0.0725 & -- \\
Incumbent: image anchor only & 0.9274 & 0.0680 & -- \\
Incumbent: text gating only & 0.9346 & 0.0615 & 13 \\
Incumbent: full \asg{} ($g_{\rm hi}{=}3.0$) & 0.9380 & 0.0586 & 8 \\
\bottomrule
\end{tabular*}
\end{table}

\begin{table}[t]
\centering
\caption{Dose selection and spatial controls.
Panels (a,b) use the MagicBrush development split; new catastrophic
turns are counted relative to text-only gating ($g_{\rm hi}=1.5$),
and (b) uses the 311 turns with source--target CLIP-I $\leq 0.97$.
Panel (c) reports PIE-Bench++ controls relative to \asg{}.}
\label{tab:dosefull}

\begingroup
\fontsize{9}{10}\selectfont
\setlength{\tabcolsep}{1.5pt}
\renewcommand{\arraystretch}{1.05}

\begin{tabular*}{\columnwidth}
{@{\extracolsep{\fill}}lccc@{}}
\toprule

\multicolumn{4}{@{}l@{}}
{\textit{(a) Anchor dose: MagicBrush (528 turns)}} \\
\midrule
$g_{\rm hi}$
& CLIP-I$\uparrow$
& L1$\downarrow$
& New cat.$\downarrow$ \\
\midrule
1.50 & 0.9346 & 0.0615 & 0 \\
2.25 & 0.9367 & 0.0598 & 0 \\
\textbf{3.00} & \textbf{0.9380} & \textbf{0.0586} & \textbf{0} \\
4.50 & 0.9350 & 0.0592 & 3 \\

\midrule
\multicolumn{4}{@{}l@{}}
{\textit{(b) ASG minus reference (311 turns)}} \\
\midrule
Reference
& $\Delta$CLIP-I$\uparrow$
& $\Delta$L1$\downarrow$
& $N$ \\
\midrule
Uniform CFG
& $+0.0147$
& $-0.0145$
& 311 \\
Text-only gating
& $+0.0031$
& $-0.0029$
& 311 \\

\midrule
\multicolumn{4}{@{}l@{}}
{\textit{(c) Control minus ASG: PIE-Bench++}} \\
\midrule
Control
& $\Delta$CLIP-W$\uparrow$
& $\Delta$CLIP-E$\uparrow$
& $\Delta$PSNR-BG$\uparrow$ \\
\midrule
Fixed guidance
& $-0.273$
& $-0.326$
& $+0.27$ \\
Per-image mean
& $-0.726$
& $-0.727$
& $+0.06$ \\

\bottomrule
\end{tabular*}

\endgroup
\end{table}

\subsection{Where the Gain Comes From}

In the probe behind Fig.~\ref{fig:discovery}, no concentration statistic separates successes from catastrophic failures --- top-10\% attention mass stays within 0.13--0.16 across the two outcomes --- so the support contributes spatial gating rather than a success signal. Aggregate similarity partly rewards doing nothing: 217 of 528 development targets (41.1\%) already exceed CLIP-I 0.97 with their source. Table~\ref{tab:dosefull}(b) shows the gain is not confined to these: on the remaining 311 turns, \asg{} improves both metrics over uniform sampling and text-only gating. Table~\ref{tab:ablation} isolates the two branches. On both checkpoints, text gating alone accounts for most of the improvement; on the incumbent, the image anchor alone improves on uniform sampling but clearly trails text gating, and added on top of text gating it yields the full method and reduces the tail from 13 to 8.

Two non-spatial controls on the full 700-image PIE-Bench++ population separate spatial allocation from guidance strength (Table~\ref{tab:dosefull}(c)). Collapsing both gates to the strongest development-split fixed dose costs both CLIP scores, and its $0.27$\,dB background gain is consistent with under-editing. Replacing each spatial map with its per-example scalar mean keeps the attention-derived dose exactly and still loses up to $0.73$ CLIP: with dose matched and only placement removed, the spatial allocation itself carries the gain, and background PSNR changes by only $0.06$\,dB, so flattening the map provides little preservation benefit. The dose sweep (Table~\ref{tab:dosefull}(a)) bounds the anchor dose: raising $g_{\rm hi}$ from 1.5 to 3.0 improves both metrics; at 4.5, CLIP-I falls, three new catastrophic turns appear, and all three are removal edits, consistent with strong anchoring resisting requested deletions.

\begin{figure}[!t]
\centering
\includegraphics[width=0.91\columnwidth]{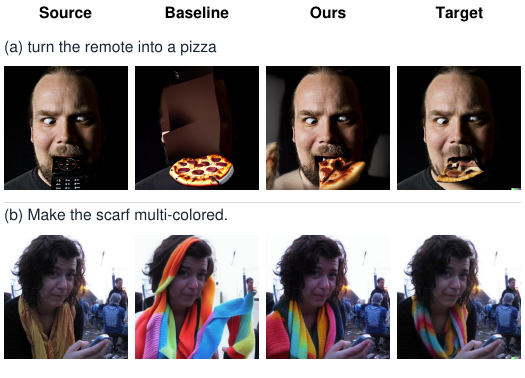}
\caption{Selected same-seed development comparisons against uniform CFG $(7.5,1.5)$. Our outputs replace the remote with pizza while retaining facial appearance (a), and recolor the scarf while better retaining its placement (b). Target images provide reference edits.}
\label{fig:evidence-strip}
\end{figure}

Fig.~\ref{fig:evidence-strip} shows matched-seed development pairs against uniform CFG: the uniform baseline drags unedited content along with the edit, while \asg{} confines the change to the instructed object, with the face recognizable and the scarf at the neck.

\subsection{Practical Scope}

The support is read from activations that the IP2P UNet already produces, and all three noise predictions in Eq.~\eqref{eq:cfg} are reused, so the wrapper retains the incumbent's 1.9-second observed latency on one A100-SXM4. The wrapper stores a single accumulated support map at half the latent resolution and resizes the two gates to the latent grid; the same decomposition extends directly to editors with separate image and text CFG branches. One configuration serves both benchmarks: no weight, threshold, or dose differs between MagicBrush and PIE-Bench++. Given the generation seed, the wrapper is deterministic, so fixed-seed comparisons are exactly reproducible.

\section{Conclusion}
We presented \method{} (\asg{}), a training-free sampler wrapper that makes the dual-CFG weights of instruction-guided editing spatial: a soft support read from the editor's own instruction attention weakens the text-edit direction outside the instructed region and strengthens the image anchor inside it, within the original sampling trajectory and with no additional network evaluations. On the full MagicBrush and PIE-Bench++ splits, \asg{} leads three of four MagicBrush metrics and PIE-Bench++ background PSNR against seven released editors, and a dose-matched control attributes the gain to the spatial allocation itself. The construction assumes an editor with separate text and image CFG branches and reads the support at the resolution of the editor's own cross-attention, with a single fixed configuration serving both benchmarks. Extending the allocation rule to newer editing backbones and adapting the anchor dose per edit are left for future work.

\bibliographystyle{IEEEbib}
\bibliography{references}
\end{document}